\pdfoutput=1

\documentclass[11pt]{article}

\usepackage[]{ACL2023}

\usepackage{times}
\usepackage{latexsym}

\usepackage[T1]{fontenc}

\usepackage[utf8]{inputenc}

\usepackage{microtype}

\usepackage{inconsolata}

\usepackage{booktabs,tabularx,caption}
\usepackage{subcaption}
\usepackage{enumitem}

\usepackage{hhline}

\newcommand{\SVAMPSym}{SVAMP-Sym}
\newcommand{\wxyz}{\texttt{(w,x,y,z)}}
\newcommand{\ijkl}{\texttt{(i,j,k,l)}}
\newcommand{\pqrs}{\texttt{(p,q,r,s)}}

\usepackage{tikz}
\usepackage{graphicx}
\usepackage{array}
\usetikzlibrary{shapes.geometric, arrows}
\tikzstyle{rect} = [rectangle, rounded corners, minimum width=3cm, minimum height=1cm,text centered, draw=black]
\tikzstyle{rect1} = [rectangle, minimum width=3cm, minimum height=1cm,text centered, draw=black]
\tikzstyle{process} = [rectangle, minimum width=3cm, minimum height=1cm, text centered, draw=black]
\tikzstyle{decision} = [diamond, minimum width=3cm, minimum height=1cm, text centered, draw=black]
\tikzstyle{arrow} = [thick,->,>=stealth]
\tikzstyle{trap} = [trapezium, trapezium left angle=80, trapezium right angle=100, minimum height=1cm, minimum width=3cm, text centered, draw=black, fill=blue!10]
\tikzstyle{dia} = [diamond, minimum width=3cm, minimum height=0.5cm, text centered, draw=black, fill=green!10]
\tikzstyle{dia1} = [diamond, minimum width=3cm, minimum height=0.5cm, text centered, draw=black, fill=red!10]

\newcommand*\colourcheck[1]{%
  \expandafter\newcommand\csname #1check\endcsname{\textcolor{#1}{\ding{52}}}%
}

\newcounter{daggerfootnote}

\usepackage{amsmath,amsfonts,bm,amssymb}
\usepackage{amsthm}
\usepackage{mathtools}
\usepackage{xcolor}
\usepackage{nicefrac}       %
\usepackage{multirow}
\usepackage{longtable}
\usepackage{parskip}

\usepackage{pythonhighlight}
\usepackage{xcolor,pifont}

\newenvironment{itemize*}%
{\begin{itemize}[leftmargin=*,topsep=0pt]%
		\setlength{\itemsep}{0pt}%
		\setlength{\parskip}{0pt}}%
	{\end{itemize}}
\newenvironment{enumerate*}%
{\begin{enumerate}[leftmargin=*,topsep=0pt]%
		\setlength{\itemsep}{0pt}%
		\setlength{\parskip}{0pt}}%
	{\end{enumerate}}

\def\eqref#1{equation~\ref{#1}}

\def\1{\bm{1}}

\DeclareMathAlphabet{\mathsfit}{\encodingdefault}{\sfdefault}{m}{sl}
\SetMathAlphabet{\mathsfit}{bold}{\encodingdefault}{\sfdefault}{bx}{n}

\usepackage[noabbrev,capitalize,nameinlink]{cleveref}
\allowdisplaybreaks

\usepackage{thmtools, thm-restate}

\title{Reasoning in Large Language Models Through\\Symbolic Math Word Problems\\\vspace{0.1in}\normalfont{\small(Appears in Findings of ACL 2023. First submitted on January 20, 2023)}}

\author{Vedant Gaur\thanks{~~Currently attending the University of Pennsylvania, although all work was done while at Aragon High School.} \\
  Aragon High School \\
  \texttt{vedantgaur101@gmail.com} \\\And
  Nikunj Saunshi\thanks{~~Most of the work was performed while at Princeton University and after graduating, but before joining Google.} \\
  Google Research, New York \\
  \texttt{nsaunshi@google.com} \\
  }

\begin{document}
\maketitle

\begin{abstract}
    \looseness-1Large language models (LLMs) have revolutionized NLP by solving downstream tasks with little to no labeled data.
    Despite their versatile abilities, the larger question of their ability to reason remains ill-understood.
    This paper addresses reasoning in math word problems (MWPs) by studying symbolic versions of the numeric problems, since a symbolic expression is a ``concise explanation'' of the numeric answer.
    We create and use a symbolic version of the SVAMP dataset and find that GPT-3's davinci-002 model also has good zero-shot accuracy on symbolic MWPs.
    To evaluate the faithfulness of the model's reasoning, we go beyond accuracy and additionally evaluate the {\em alignment} between the final answer and the outputted reasoning, which correspond to numeric and symbolic answers respectively for MWPs.
    We explore a {\em self-prompting} approach to encourage the symbolic reasoning to align with the numeric answer, thus equipping the LLM with the ability to provide a {\em concise and verifiable} reasoning and making it more interpretable. Surprisingly, self-prompting also improves the symbolic accuracy to be higher than both the numeric and symbolic accuracies, thus providing an ensembling effect.
    The {\SVAMPSym} dataset will be released for future research on symbolic math problems.
    
\end{abstract}

\section{Introduction}
\label{sec:intro}

\looseness-1Large language models (LLMs), with hundreds of billions of parameters, can solve a wide range of NLP tasks such as machine translation, question-answering, etc., taking us closer to general-purpose intelligent agents.
The initial success of GPT-3 \citep{brown2020language} has led to many other LLMs~\citep{rae2021scaling,smith2022using,chowdhery2022palm} which have, perhaps surprisingly, taken big strides in solving difficult tasks like common sense reasoning, math and science problems \citep{lewkowycz2022solving}, and writing code \citep{li2022competition}.

\looseness-1Despite the incredible successes, we have little understanding of why LLMs are effective at problems that require reasoning. 
In fact we have limited techniques to quantifiably study the question of reasoning beyond just evaluating accuracy.
Recent ideas like Chain-of-Thought prompting (CoT) \citep{wei2022chain,kojima2022large} encourage the model to ``think step by step'' and output a verbose reasoning in text.
However, verifying such reasoning at scale will incur the infeasible cost of manually going over the text outputs.
Furthermore, we would like the model's reasoning to be consistent with its outputted answer, in order to trust the presented reasoning.
For these considerations, we would like our models to output a {\em concise reasoning} or explanation for its answer that can be {\em automatically verified}.
In particular, we desire reasoning in the form of explanations that are
\begin{itemize}
    \item Verifiable: For ease of evaluating correctness of the outputted reasoning, and
    \item Concise: For scalability of verification. Manually going through text reasoning can quickly get cumbersome
\end{itemize}
For instance, instead of a text description of an algorithm to solve a problem, a Python implementation of the algorithm would be a more concise explanation for the reasoning behind the algorithm\footnote{We can automatically verify the answer not just for one problem, but for all instance of that problem}.
Similarly, a simple linear model or decision tree explaining the answers of a black-box neural network also achieves the same goal \citep{ribeiro2016should}.
Concise explanations can provide clearer insights into the reasoning abilities of models, and verifiable explanations aid interpretability and help foster trust in models, in line with explainable AI \citep{samek2019explainable}.

\begin{figure*}[t!]
    \centering
    \includegraphics[scale=0.28]{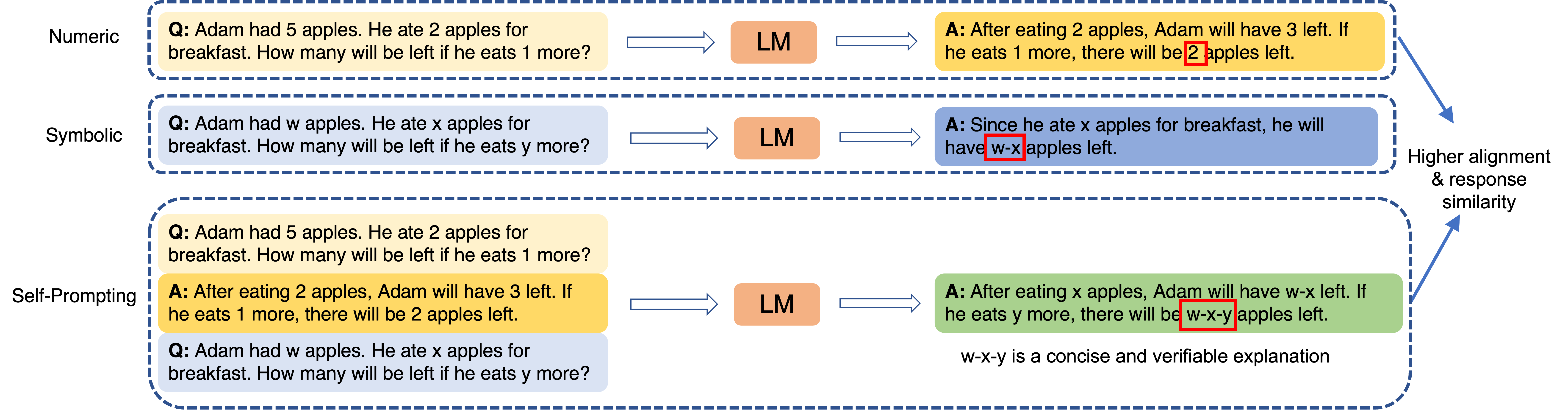}
    \caption{LMs can be queried to solve numeric/symbolic math problems. Self-prompting includes the numeric problem and the LM's solution to it before passing the symbolic problem. This encourages the model to output the answer that aligns with the numeric answer. The symbolic expression \texttt{w-x-y} serves as a concise explanation/reasoning for the numeric answer of 2.\vspace{-0.1in}}
    \label{fig:methods}
\end{figure*}

\looseness-1In this work we use concise and verifiable explanations to study reasoning abilities of LLMs in math word problems (MWPs).
LLMs have shown to achieve good zero-shot accuracy on many numeric MWP benchmarks \citep{kojima2022large}.
Chain-of-thought like ideas encourage LLMs to first general a step-by-step explanation (in text) before generating the answer.
However, this does not satisfy the criteria of being concise or easily verifiable\footnote{It is not uncommon for the outputted reasoning to be inconsistent with the final answer}.
We address reasoning by considering symbolic versions of numeric MWPs, because a symbolic expression can be viewed as a concise explanation for a numeric answer and can also be automatically evaluated.
Thus in this reasoning framework for MWPs, we require an LLM to output both, a numeric answer and a concise symbolic expression, such that we have: (1) high accuracy for the predicted numeric answer, (2) high alignment of the symbolic expression with the predicted numeric answer.
While most prior studies focus on goal (1), we argue that goal (2) is equally important for interpretability of these models and to trust the its reasoning. 
Our main finding is that LLMs can also do reasonably well on goal (2), either by generating a numeric answer and symbolic explanation together, or by generating the answer first and then a post-hoc symbolic explanation.
In this context, we make the following contributions:

\looseness-1\noindent\textbf{Symbolic evaluation.} We construct a symbolic version of the SVAMP dataset \citep{patel2021nlp} called {\SVAMPSym} to evaluate LLMs.
Firstly we find, perhaps surprisingly, that GPT-3’s davinci-002 model already achieves good zero-shot accuracy on symbolic problems ($64.2$\%), comparable to the numeric accuracy of $68.9$\%.
Secondly, this observation provides a simple way to get good accuracy and alignment for numeric problems by first solving symbolic versions and then substituting back the values for variables.
This approach generates the numeric answer and a symbolic explanation in one go, thus trivially achieving\footnote{\noindent If a ``calculator'' can evaluate symbolic expressions.} an accuracy of $64.2$\% and alignment of $100$\%.

\looseness-1\noindent\textbf{Self-prompting.} There are two key drawbacks with the above approach: (a) symbolic accuracy of $64.2$\% is lower than the numeric accuracy ($68.9$\%), (b) alignment of symbolic expressions, as post-hoc explanation to the original numeric answers, is very low ($\sim50$\%).
To get a better post-hoc explanation, we propose a novel {\em self-prompting} approach that first prompts the LLM with the numeric problem and its response to the problem, and then asks it to solve the symbolic problem; see \Cref{fig:methods}.
Self-prompting significantly improves alignment with numeric answers to $74$\% (a $24$\% absolute improvement).
Surprisingly, self-prompting also improves the symbolic accuracy to $71.7$\%, higher than both the raw numeric and symbolic accuracies of $68.9$\% and $64.2$\% respectively. This suggests that self-prompting has an ensembling effect.

\looseness-1We perform further ablation studies and analyses and hope that these insights will aid future work on using LLMs for reasoning problems.

\subsection{Related Work}
\label{sec:prior}

\looseness-1Language models like GPT-3 \citep{brown2020language} and MLMs like BERT \citep{devlin2019bert} have demonstrated impressive emergent behaviors \citep{wei2022emergent} at scale.
For math problems, Minerva \citep{lewkowycz2022solving} was fine-tuned from PaLM \citep{chowdhery2022palm} to do well on many MWP benchmarks.
Instead of fine-tuning, \citet{wei2022chain} uses in-context learning and finds that asking the model to ``think step by step'' (CoT prompting) improves few-shot accuracy on MWPs; \citet{kojima2022large} verify this for zero-shot setting as well, which is the focus of our work.

\looseness-1There is limited theoretical work for the downstream success of LMs \citep{saunshi2021a,xie2022an} and the emergent behaviors of LLMs through scaling laws \citep{kaplan2020scaling}.
Our idea of self-prompting is motivated by the efficacy of in-context learning \citep{brown2020language} and prompting \citep{liu2023pre} in LMs.
The ensembling effect of self-prompting idea could be related to self-calibration abilities of LMs \citep{kadavath2022language}.
Finally, \citet{ho2022large} survey the progress of LMs on various notions of reasoning; we consider a weaker notion of ``concise post-hoc explanations'' here.

\section{Math Word Problems with LLMs}
\label{sec:llm_math}

\subsection{{\SVAMPSym} Dataset}
\label{sec:dataset}

\looseness-1We choose the SVAMP dataset \citep{patel2021nlp} for testing LMs on MWPs because it provides numeric answers in the form of numeric expressions (rather than just numeric values).
This lets us automatically convert the dataset into a symbolized version, without manual annotation.
The main idea is to replace all occurrences of numbers in the problem statement with newly introduced variables, e.g. {\wxyz}.
\Cref{sec:symbolized} provides more details on the dataset construction.
The dataset is released in \url{https://github.com/vedantgaur/Symbolic-MWP-Reasoning}.

\subsection{Querying and Evaluating LMs}
\label{sec:query_eval}

\looseness-1Broadly, our evaluation pipeline has four phases: (1) get a verbose response from the LLM for the math problem, (2) prompt the LLM to extract just the answer (number or symbolic expression) from its initial response, (3) refine the extracted answer using a novel {\em filtering} step, (4) compare the filtered answer to the ground-truth answer.

\looseness-1\noindent\textbf{Initial response.}
We query the LM with the problem statement and an optional CoT prompt, i.e. \texttt{"Q: <Problem> A:"} or \texttt{"Q: <Problem> A: Let's think step by step."}.
\texttt{<Problem>} could either be a numeric or symbolic problem.
\Cref{table:prompt_format} summarizes the prompts used for various settings.

\looseness-1\noindent\textbf{Answer extraction.}
Since the LLM outputs a long text response (\Cref{fig:methods}), we use an extraction prompt to isolate the answer, similar to \citet{kojima2022large}.
We query the LM with the transcript so far, followed by the question and the prompt \texttt{"The final answer (only the number) is:"} to isolate the numeric answer.
\Cref{table:prompt_format} has the similar prompt for symbolic problems.

\looseness-1\noindent\textbf{Answer filtering.}
The extraction prompt does not always isolate the final answer and sometimes outputs a sentence, especially for symbolic problems.
Thus we add a LM-independent filtering step which includes stripping escape sequences, removing commas, de-latexifying equations, picking the longest symbolic expression, among others; more details in \Cref{sec:filtering}.

\looseness-1\noindent\textbf{Answer evaluation.}
We compare the filtered answer to the ground-truth answer (symbolized expression or numeric value).
Since there are multiple ways to express the same symbolic expression (e.g. \texttt{"w + (y + x)"} and \texttt{"w + x + y"}), we compare two expressions through their evaluations on 20 random variable assignments.
If they match on all 20 assignments, we adjudge them to be equivalent, making a (reasonable) assumption that 20 random assignments will avoid false positives.

\begin{table*}[t!]
    \begin{center}
        
    \begin{tabular}{cc|c|ccccc|}
         \cline{3-8}
         &  & \text{Numeric} & \multicolumn{5}{c|}{\text{Symbolic}} \\

         &  & \textbf{} & \multicolumn{3}{c|}{\small\wxyz} & \multicolumn{1}{c|}{\small \pqrs} & {\small \ijkl} \\

         \cline{2-8}
         \multicolumn{1}{c|}{} & {Evaluation} & Raw {\scriptsize (-F)} & \multicolumn{1}{c|}{{Raw {\scriptsize (-F)}}} & \multicolumn{1}{c|}{{\textbf{SP} {\scriptsize (-F)}}} & \multicolumn{1}{c|}{\textbf{SP + AP}} & \multicolumn{1}{c|}{Raw} & \multicolumn{1}{c|}{Raw} \\

        \hhline{|=|=|=|=|=|=|=|=|}
         \multicolumn{1}{|c|}{\multirow{2}{*}{{Accuracy}}} & \textit{Vanilla} & 65.6 {\scriptsize(61.6)} & \multicolumn{1}{c|}{59.7 {\scriptsize(47.6)}} & \multicolumn{1}{c|}{61.9 {\scriptsize(40)}} & \multicolumn{1}{c|}{\textbf{68.3}} & \multicolumn{1}{c|}{62.3} & 53.5 \\

         \cline{2-8} 
         \multicolumn{1}{|c|}{} & \textit{CoT} & 68.9 {\scriptsize(65.9)} & \multicolumn{1}{c|}{64.2 {\scriptsize(48.8)}} & \multicolumn{1}{c|}{67.9 {\scriptsize(48.6)}} & \multicolumn{1}{c|}{\textbf{71.7}} & \multicolumn{1}{c|}{64.4} & 58.4 \\ 
         
         \cline{1-8}
         \multicolumn{1}{|c|}{\multirow{2}{*}{{Alignment}}} & \textit{Vanilla} & - & \multicolumn{1}{c|}{52.9 {\scriptsize(40.7)}} & \multicolumn{1}{c|}{60.3 {\scriptsize(40)}} & \multicolumn{1}{c|}{\textbf{64.9}} & \multicolumn{1}{c|}{56.3} & 44.7 \\ 
         
         \cline{2-8} 
        \multicolumn{1}{|c|}{} & \textit{CoT} & - & \multicolumn{1}{c|}{51.2 {\scriptsize(39.1)}} & \multicolumn{1}{c|}{63.1 {\scriptsize(44.9)}} & \multicolumn{1}{c|}{\textbf{74}} & \multicolumn{1}{c|}{51.9} & 47.1 \\

        \hhline{|=|=|=|=|=|=|=|=|}
        \multicolumn{1}{|c|}{Similarity} & \textit{Vanilla} & - & \multicolumn{1}{c|}{27.8} & \multicolumn{1}{c|}{44.2} & \multicolumn{1}{c|}{\textbf{49.8}} & \multicolumn{1}{c|}{27.1} & 26.8 \\ 
        
        \cline{2-8} 
        \multicolumn{1}{|c|}{\small (BLEU)} & \textit{CoT} & - & \multicolumn{1}{c|}{21.3} & \multicolumn{1}{c|}{53.9} & \multicolumn{1}{c|}{\textbf{57.6}} & \multicolumn{1}{c|}{22.7} & 21.4 \\ 
        
        \cline{1-8} 
        \multicolumn{1}{|c|}{Similarity} & \textit{Vanilla} & - & \multicolumn{1}{c|}{56.5} & \multicolumn{1}{c|}{65.2} & \multicolumn{1}{c|}{\textbf{71.3}} & \multicolumn{1}{c|}{56.8} & 55.4 \\
        
        \cline{2-8} 
        \multicolumn{1}{|c|}{\small (Levenshtein)} & \textit{CoT} & - & \multicolumn{1}{c|}{44.9} & \multicolumn{1}{c|}{75.6} & \multicolumn{1}{c|}{\textbf{79.8}} & \multicolumn{1}{c|}{45.4} & 43.9 \\ \hline
    \end{tabular}
    \caption{Zero-shot accuracy and alignment evaluations using GPT-3. All values are reported in \%.
    ``Raw'' refers to evaluation on the SVAMP and ({\SVAMPSym}) dataset for numeric (symbolic) MWPs; (-F) refers to the output before the filtering step. ``SP'' is the new self-prompting method and ``SP + AP'' refers to two-stage self-prompting where we an additional ``Alignment Prompt'' is added when needed; see \cref{sec:self_prompting}. 
    CoT prompting consistently elicits higher accuracy from the model for numeric and symbolic problems.
    While accuracy and alignment only look at the final answers, we also measure similarity between the full responses for numeric and symbolic problems. As evident, self-prompting significantly improves the similarity under BLEU score and Levenshtein metric; \Cref{sec:similarity} has more details on these metrics.\vspace{-0.2in}}
    \label{table:main_data}
    \end{center}
\end{table*}

\section{Experimental Results}
\label{sec:results}

\looseness-1We pick 150/1000 examples from the SVAMP dataset (due to budget constraints) and run each examples 5 times.
We use GPT-3's davinci-002 model with temperature 0.0 for (mostly) deterministic outputs, with a max token length of 256.

\subsection{Numeric and Symbolic Evaluations}
\label{sec:numeric_symbolic_exp}

We  discuss the accuracies for solving numeric and symbolic math problems from SVAMP and {\SVAMPSym} respectively.

\looseness-1\noindent\textbf{Numeric accuracy.}
The zero-shot numeric accuracy both with chain-of-thought (CoT) prompt and without (vanilla) are presented in \Cref{table:main_data}; they are $68.9$\% and $65.6$\% respectively. 
This good performance is unsurprising given prior work \citep{kojima2022large}.
Our accuracies are $\sim5$-$7$\% higher than \citet{kojima2022large}, due in part to better answer extraction and filtering.

\looseness-1\noindent\textbf{Symbolic accuracy.}
We also evaluate raw symbolic problems from {\SVAMPSym} in the vanilla and CoT settings with 3 natural choices for variables: {\wxyz}, {\ijkl} and {\pqrs}.
Firstly we observe, in \Cref{table:main_data}, that GPT-3 can achieve pretty high symbolic accuracies with variables {\wxyz}: vanilla and CoT settings achieve $59.7$\% and $64.2$\% respectively, which is just $4$-$5$\% lower than numeric accuracy.
Furthermore, we notice that variables {\ijkl} have slightly worse accuracy than other variable settings, possibly because {\wxyz} and {\pqrs} are more popular choice for variables in the training data for language models.

\looseness-1\noindent\textbf{Effect of filtering.}
We report the accuracies without the filtering step in \Cref{table:main_data}; these are the (-F) entries.
While there is a $4$-$5$\% drop in the numeric accuracy without filtering, the drop is $12$-$14$\% for symbolic problems, suggesting that filtering is much more crucial for symbolic problems\footnote{Intuitively it makes sense that extracting an expression/equation is harder than extracting a single number}. 
Our extraction and filtering steps still have issues and there is scope for improvement.

\subsection{Reasoning and Alignment}
\label{sec:reasoning_alignment}
\looseness-1While prior work only cares about the accuracy on MWPs, we also study of reasoning abilities of LLMs by requiring them to generate a concise explanation for numeric answers in the form of a symbolic expressions.
We evaluate ``reasoning ability'' through an alignment metric that checks if the outputted numeric answer and symbolic expression compute to the same value.
In general there is no consistent zero-shot method to return a perfectly aligned symbolic expression.
A natural attempt to generate such an expression is to directly solve the symbolic versions of numeric problem.
However this approach has very low alignment, i.e. the symbolic output does not reflect the way in which the model solved the numeric problem.
Specifically in \Cref{table:main_data}, the average alignment score for raw symbolic outputs is only $52.9\%$ and $51.2\%$ for Vanilla and CoT respectively.
This motivates self-prompting.

\subsection{Self-prompting}
\label{sec:self_prompting}
In order to improve alignment, we propose a two-step procedure that first inputs the numeric MWP and the LM's response to it, followed by the symbolic version of the MWP.
In particular the prompt looks like \texttt{"Q: <Numeric Question> A: <Model Response> Q: <Symbolic Question> A:"}.
Given the in-context tendencies of LMs, we hope that this encourages the symbolic response to imitate the numeric response and thus return a well aligned expression.
We find in \Cref{table:main_data} that this approach (termed \textbf{SP}) indeed improves the alignment by $\sim10$\% over the naive approach.

\looseness-1We take this one step further: whenever the numeric and symbolic answers do not align, we add another ``alignment prompt'' before the symbolic problem that explicitly asks the model to copy the numeric answer; see \Cref{table:prompt_format} for the exact format.
Results in the \textbf{SP+AP} column of \Cref{table:main_data} verify that this leads to another $11$\% improvement over \textbf{SP} and $\sim22$\% improvement over raw symbolic.
Surprisingly we find that \textbf{SP+AP} has higher accuracy than raw numeric and raw symbolic, suggesting a ``best of both worlds'' or ensembling phenomenon in action.
Further analysis in \Cref{fig:tag_barplot} reveals how self-prompting combines the benefits of numeric and symbolic.

\looseness-1We also compute the similarity between the full numeric and symbolic responses. \Cref{table:main_data} reveals that the average similarity is significantly higher for \textbf{SP} and \textbf{SP+AP} compared to raw symbolic. So not only do the answers align more but also the full text responses are very similar.
Histograms of similarity scores can be found in \Cref{sec:similarity}.
Additional analyses and results can be found in \Cref{sec:ablations_analysis}.

\section{Conclusions and Future Work}
\label{sec:conclusion}

\looseness-1This paper studies reasoning in LLMs for MWPs and results suggest that LMs are good at zero-shot solving of symbolic MWPs, and that this ability can lead to concise explanations.
Self-prompting emerges as a promising idea to generate better explanations and the ensembling effect demonstrated by it can potentially have other applications (left for future work).
Alignment with self-prompting, while significantly better than with raw symbolic outputs, still has a lot of scope for improvement.
Aspects that are not considered are few-shot learning of explanations and the role of temperature, which could improve accuracy and alignment.
Finally the notion of ``concise explanation'' to study reasoning can have implications beyond MWPs.

\paragraph{Broader Impact Statement.}
Given the incredible successes of LLMs, it is becoming increasingly important to study why they work and how to debug them when they are wrong.
There are ongoing debates and discussions about whether LMs are simply ``stochastic parrots'' \citep{bender2021dangers} or they actually ``understand'' language.
Besides there are also privacy concerns \citep{carlini2021extracting} associated with LLMs trained on extremely large corpora.
Our work attempts to formalize a weak notion of ``reasoning'' in math problems that could help with improving the intepretability, and thus trustworthiness, of such models.
This is extremely important if LLMs are to be deployed in real-life applications.
That said, any preliminary notion or definition of ``reasoning in LLMs'', including the one in this paper, should be taken with a healthy dose of skepticism.

\paragraph{Acknowledgments.} We thank Misha Khodak for comments on an earlier version of this draft. We also thank the anonymous ACL reviewers for useful suggestions.

\bibliography{references}
\bibliographystyle{acl_natbib}

\appendix

\colourcheck{green}

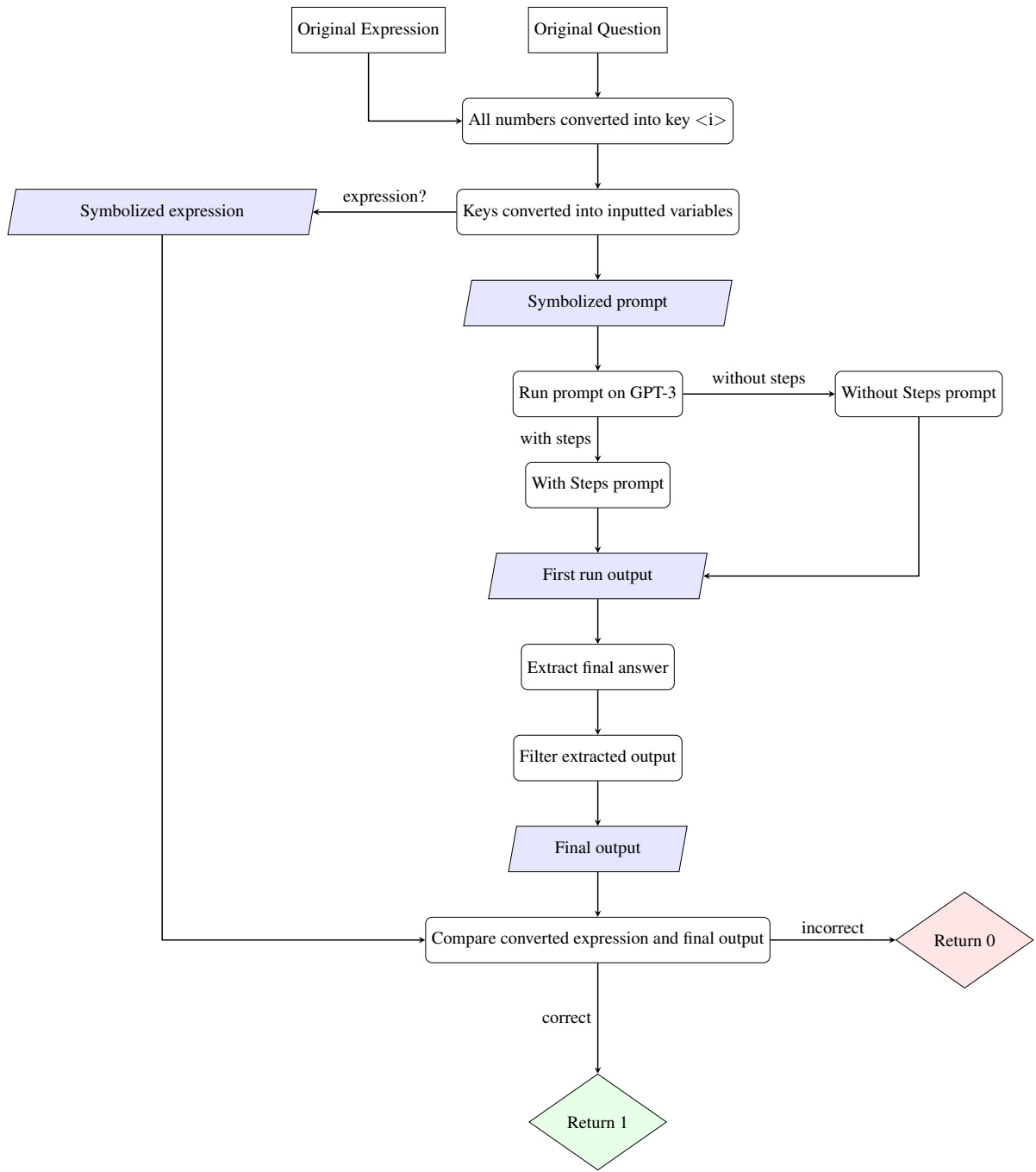
\begin{figure*}[h]

\centering
\resizebox{.7\textwidth}{!}{%
\makebox[\textwidth][c]{\begin{tikzpicture}[node distance=2cm]

\node (original) [rect1] {Original Question};
\node (keys) [rect, below of=original] {All numbers converted into key $<$i$>$};
\node (symbolize) [rect, below of=keys] {Keys converted into inputted variables};
\node (symbolized) [trap, below of=symbolize] {Symbolized prompt};
\node (gpt first) [rect, below of=symbolized] {Run prompt on GPT-3};
\node (without) [rect, right of=gpt first, xshift=5cm] {Without Steps prompt};
\node (with) [rect, below of=gpt first] {With Steps prompt};
\node (output first) [trap, below of=with] {First run output};
\node (extract) [rect, below of=output first] {Extract final answer};
\node (parse) [rect, below of=extract] {Filter extracted output};
\node (final) [trap, below of=parse] {Final output};
\node (compare) [rect, below of=final] {Compare converted expression and final output};

\node (correct) [dia, below of=compare, yshift=-2cm] {Return 1};
\node (incorrect) [dia1, right of=compare, xshift=6cm] {Return 0};

\node (expression) [rect1, right of=original, xshift=-7cm] {Original Expression};
\node (expression output) [trap, left of=symbolize, xshift=-7.5cm] {Symbolized expression};

\draw [arrow] (original) -- (keys);
\draw [arrow] (keys) -- (symbolize);
\draw [arrow] (symbolize) -- (symbolized);
\draw [arrow] (symbolized) -- (gpt first);
\draw [arrow] (gpt first) -- node[anchor=south] {without steps} (without);
\draw [arrow] (gpt first) -- node[anchor=east] {with steps} (with);
\draw [arrow] (with) -- (output first);
\draw [arrow] (without) |- (output first);
\draw [arrow] (output first) -- (extract);
\draw [arrow] (extract) -- (parse);
\draw [arrow] (parse) -- (final);
\draw [arrow] (final) -- (compare);

\draw [arrow] (expression) |- (keys);
\draw [arrow] (symbolize) -- node[anchor=south] {expression?} (expression output);
\draw [arrow] (expression output) |- (compare);

\draw [arrow] (compare) -- node[anchor=south] {incorrect} (incorrect);
\draw [arrow] (compare) -- node[anchor=east] {correct} (correct);

\end{tikzpicture}}}

\caption{Flowchart of the pipeline from an original expression to correct or incorrect outputs. The purple cells represent the outputs of the GPT-3 model as well as output processing. Both the "Original Expression" and "Original Question" at the top in rectangular cells are numeric, baseline prompts.}
\label{fig:flowchart}

\end{figure*}

\begin{figure*}[th!]
    \centering
    \includegraphics[scale=0.5]{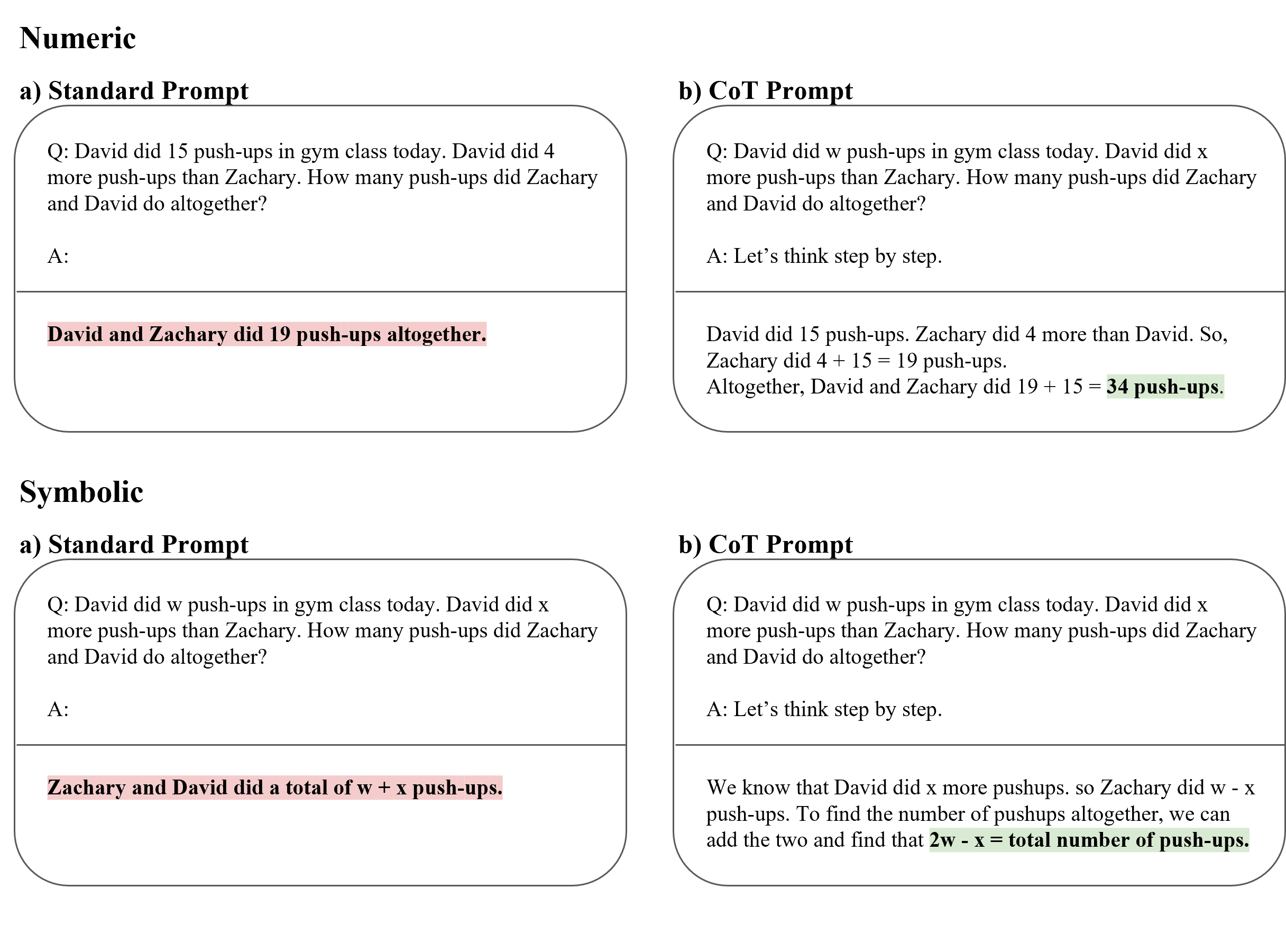}
    \caption{Examples of the input-output GPT-3 sequence of both numeric (above) and symbolic (below) runs. CoT prompting, as reported in previous papers, elicits much more detailed, and oftentimes correct outputs from the model through the additional reasoning step. We find that the use of the prompt is not exclusive to numeric reasoning, and are able to identify similar processes in symbolic runs.}
    \label{fig:svamp_exmple}
\end{figure*}

\begin{figure*}[th!]
    \centering
    \includegraphics[scale=0.85]{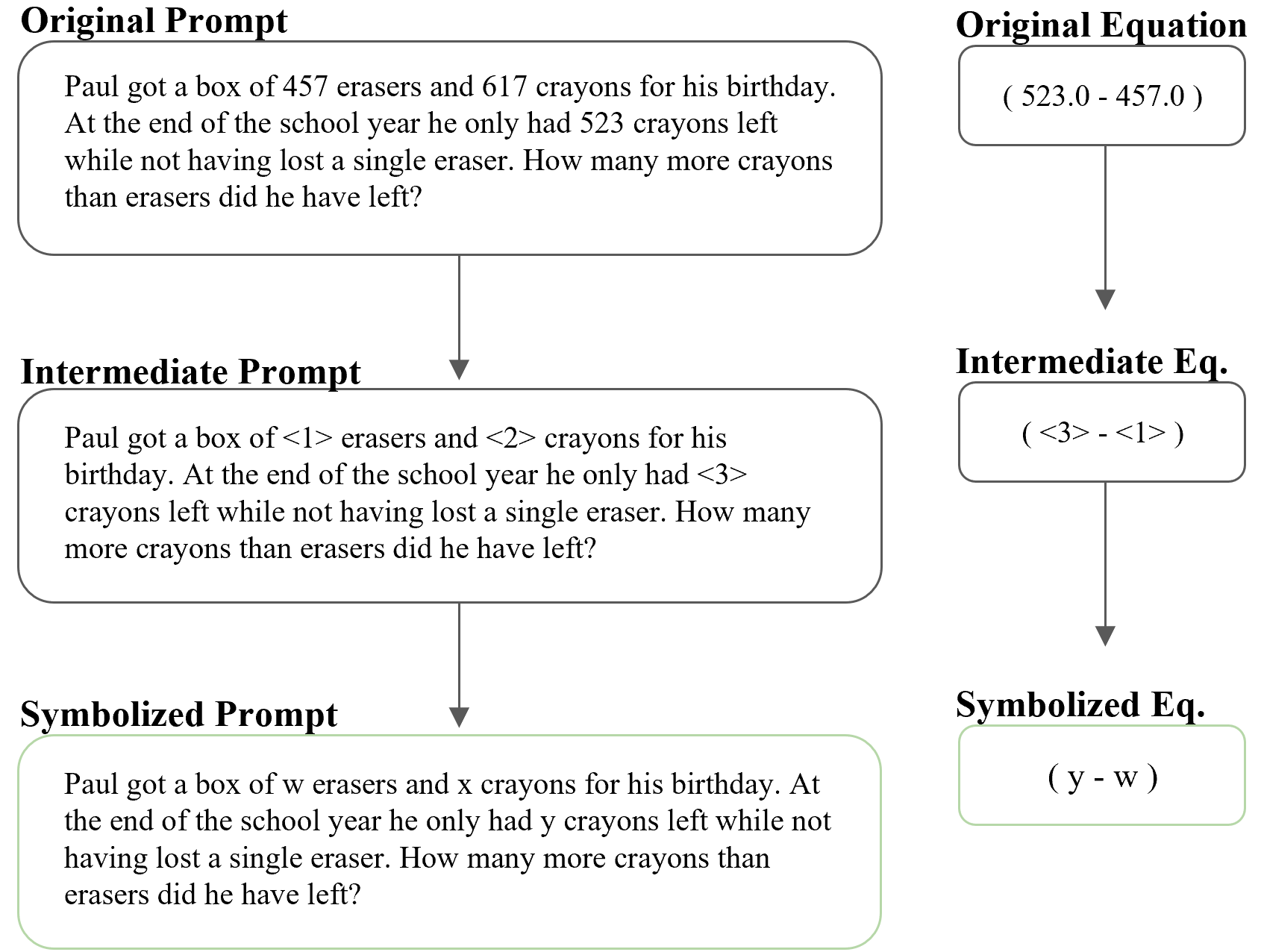}
    \caption{The process of converting a numeric problem into a symbolic one. The answer to the problem is an expression given by the SVAMP dataset, so we can easily convert it to a symbolic equation. \Cref{sec:symbolized} has more details on how this symbolization was implemented.}
    \label{fig:numeric_to_symbolic}
\end{figure*}

\section{Symbolized Dataset}
\label{sec:symbolized}

We employ a multi-step process to convert the original SVAMP \citep{patel2021nlp} prompts into its symbolic version {\SVAMPSym}, motivated by the symbolic construction in \citet{gaur2022symbolic}.
The SVAMP dataset is under the MIT License. Our {\SVAMPSym} dataset has exactly the same set of 1000 problems as SVAMP.
Given a common math word problem (MWP) from SVAMP, we parse through the given text for all numbers, which are stored in a list. Using regex, the index of the numbers can be found and replaced with keys for future replacement with variables. We use $<$i$>$ (where $i \in [1, 4]$ as there are at most four numbers in each problem definition) as keys. As shown in \Cref{fig:numeric_to_symbolic}, by generalizing the converted prompt, we allow for easy manipulation of prompts to whatever variable a user wants to use and test for downstream tasks. We then convert the keys to their respective variables. For our tests we primarily use the variables {\wxyz} for a few main reasons:
\begin{enumerate}[leftmargin=*]
    \item This set of variables is the most common in general mathematical word problems and thus makes the most sense to use as variables as opposed to an arbitrary sequence of random, or even consecutive letters.
    \item We find that the use of variables such as $x_1$, $x_2$, ..., $x_n$ (x1, x2, ..., xn when inputted into the model) many times confuses the model into conflating the simulated subscript as a coefficient.
    \item{We are able to see that the model achieves similar, if not greater accuracies with the use of {\wxyz} as opposed to other sequences of variables, see \Cref{table:main_data}}.
\end{enumerate}
Moreover, the use of a predetermined length of variables is also possible due to the aforementioned maximum number of four numbers for each prompt in the SVAMP dataset.

See \Cref{fig:numeric_to_symbolic} for an example problem, its answer, and our symbolized version of it.

\section{Ablations and Analysis}
\label{sec:ablations_analysis}

\subsection{Response Similarity}
\label{sec:similarity}

\begin{figure*}[t!]
    \centering
    \begin{subfigure}{.33\textwidth}
      \centering
      \includegraphics[width=.9\linewidth]{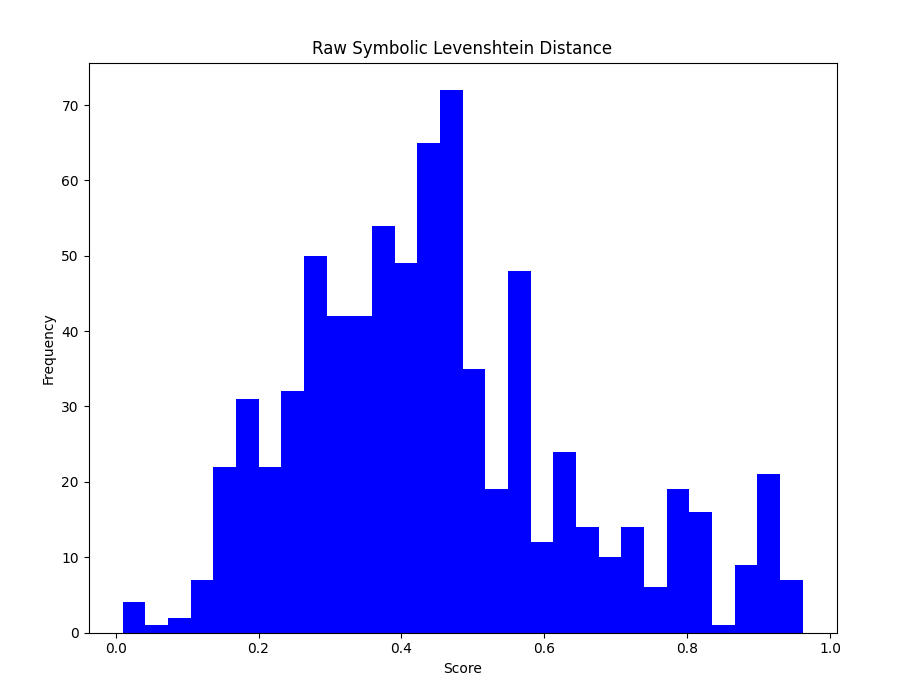}
      \caption{Raw symbolic}
      \label{fig:lev_sym}
    \end{subfigure}%
    \begin{subfigure}{.33\textwidth}
      \centering
      \includegraphics[width=.9\linewidth]{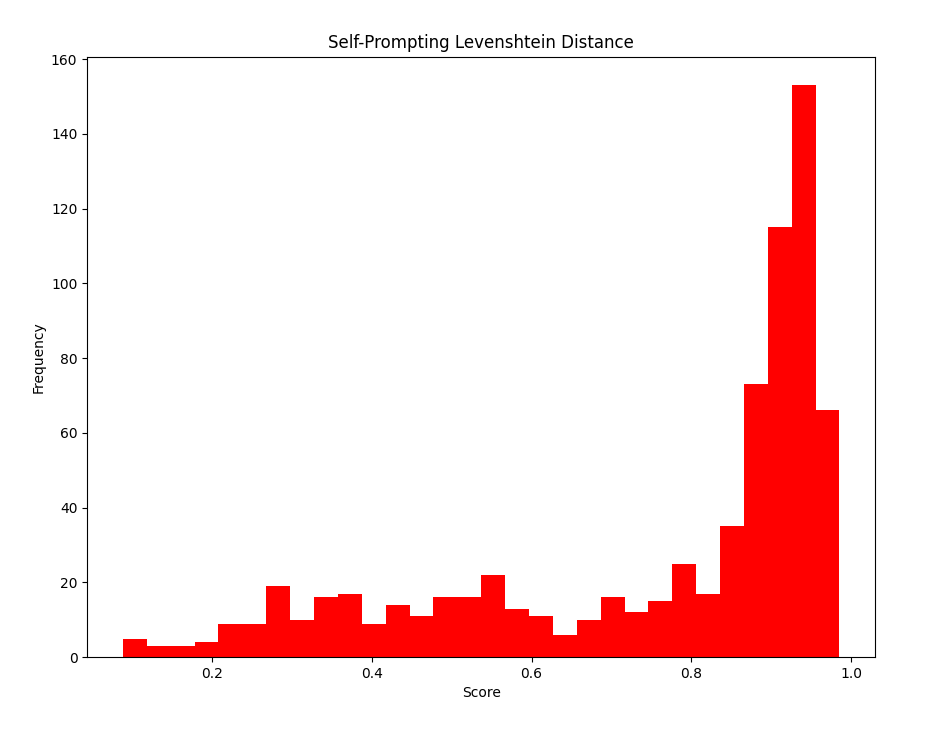}
      \caption{Self-Prompting}
      \label{fig:lev_num_sym}
    \end{subfigure}
    \begin{subfigure}{.33\textwidth}
      \centering
      \includegraphics[width=.9\linewidth]{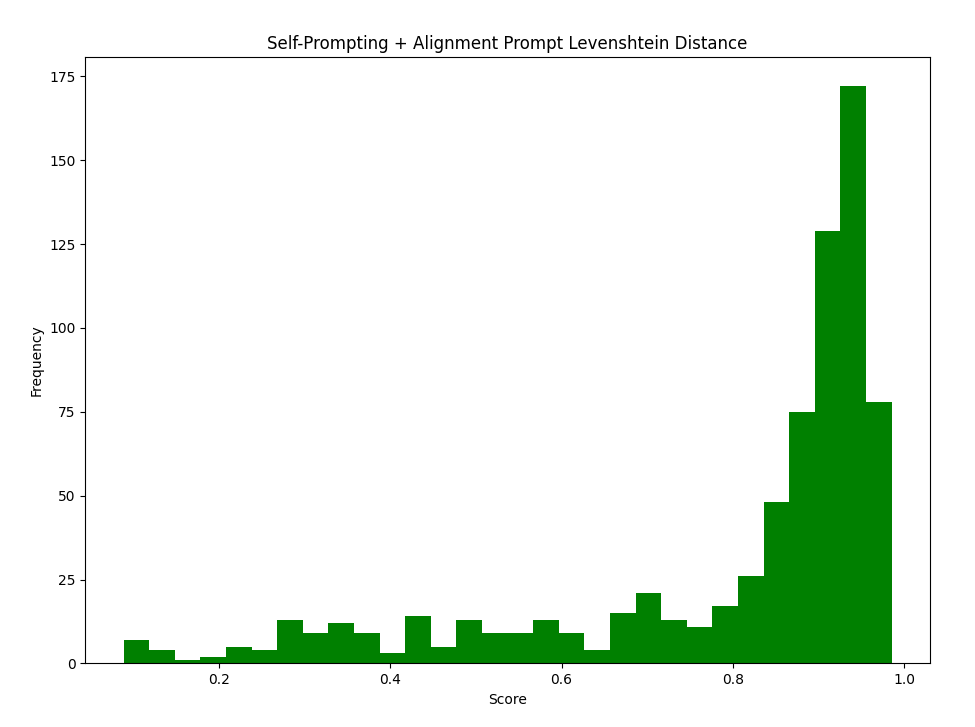}
      \caption{Self-Prompting + Alignment Prompt}
      \label{fig:lev_num_sym_plus}
    \end{subfigure}
    \caption{Levenshtein distance calculated on raw symbolic, self-prompting, and self-prompting with an additional alignment prompt outputs. Values near 1.0 (to the right) denote two sentences with very similar syntactic similarity. As evident in the graphs above, the distribution of both (b) and (c) are much more heavily skewed to the right with unimodal peaks near 1.0, whereas the distribution in (a) is shifted much more to the left. This means that both (b) and (c) are much more similar to the outputs they were compared with (numeric) than (a), highlighting the efficacy of self-prompting in mirroring numeric responses.}
    \label{fig:lev_plots}
\end{figure*}

\begin{figure*}[t!]
    \centering
    \begin{subfigure}{.33\textwidth}
      \centering
      \includegraphics[width=.9\linewidth]{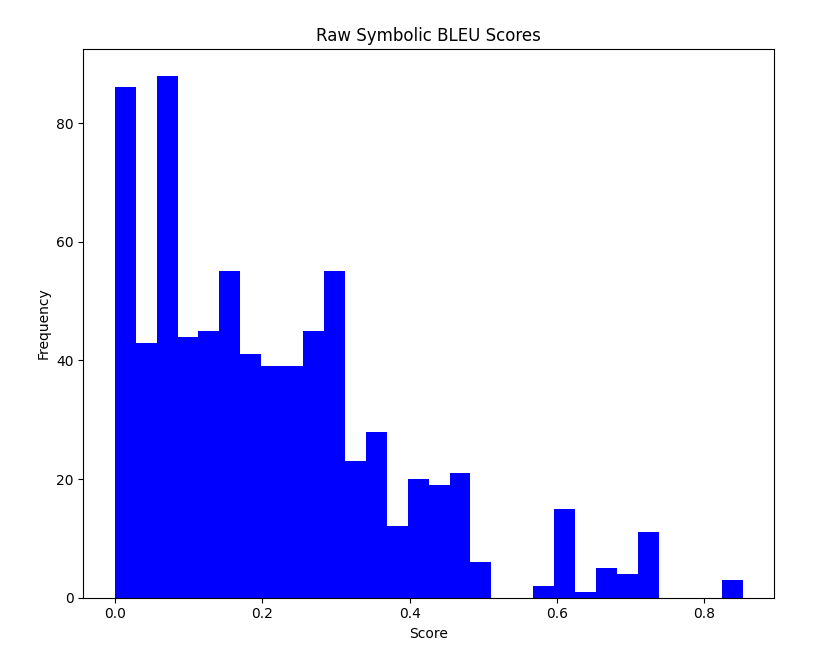}
      \caption{Raw symbolic}
      \label{fig:bleu_sym}
    \end{subfigure}%
    \begin{subfigure}{.33\textwidth}
      \centering
      \includegraphics[width=.9\linewidth]{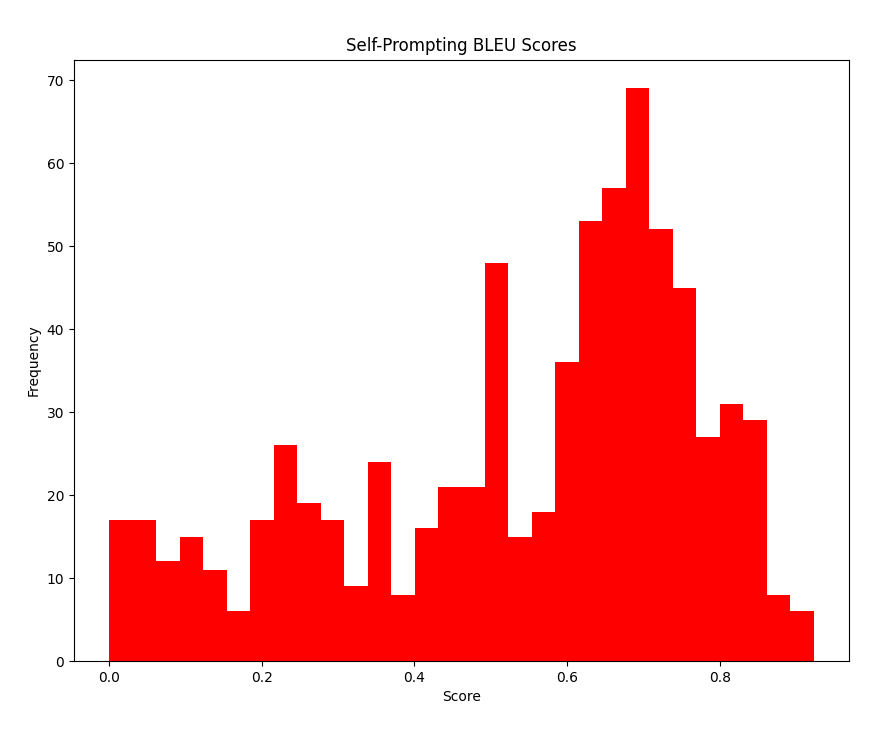}
      \caption{Self-Prompting}
      \label{fig:bleu_num_sym}
    \end{subfigure}
    \begin{subfigure}{.33\textwidth}
      \centering
      \includegraphics[width=.9\linewidth]{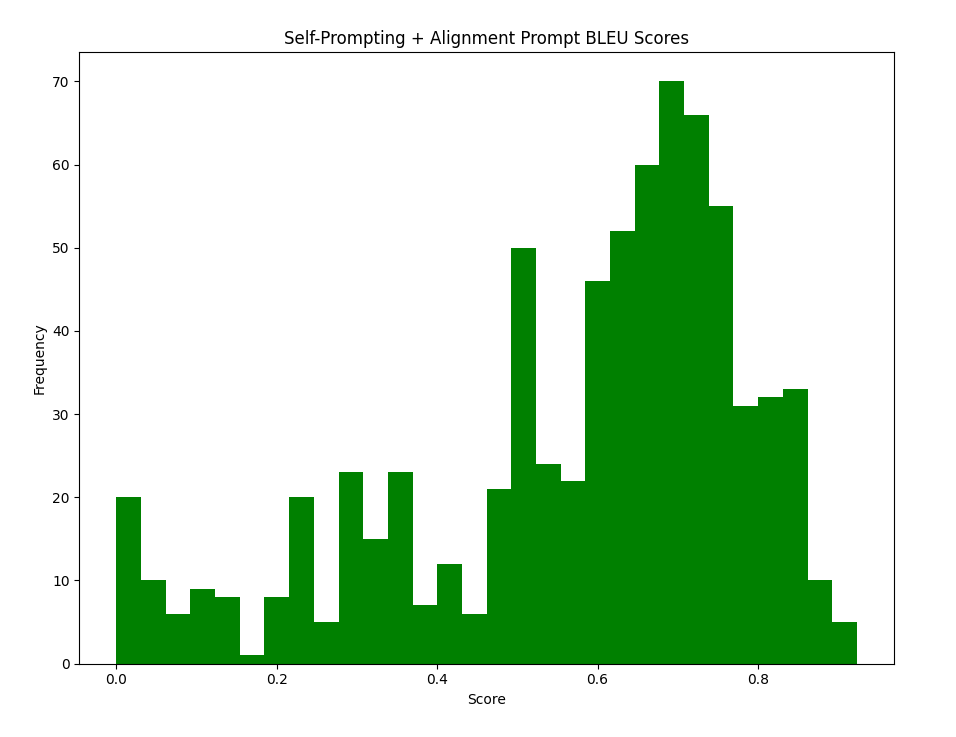}
      \caption{Self-Prompting + Alignment Prompt}
      \label{fig:bleu_num_sym_plus}
    \end{subfigure}
    \caption{BLEU Scores calculated on raw symbolic, self-prompting, and self-prompting with an additional alignment prompt outputs. As with \cref{fig:lev_plots}, values near 1.0 (to the right) denote two sentences with very similar syntactic similarity. In this instance, the BLEU Score was calculated by tokenizing and comparing the numeric outputs with respective outputs in (a), (b), and (c). This value was then normalized and plotted as described in the figures above. Both (b) and (c) both show more left-skewed distributions, while (a) models a right-skewed one. Similar to \cref{fig:lev_plots}, the use of BLEU Scores highlights how self-prompting helps with the alignment of numeric and symbolic outputs.
    }
    \label{fig:bleu_plots}
\end{figure*}

To find the syntactical similarity between the numeric and symbolic responses, we employ two main metrics: BLEU Scores and Levenshtein Distances.
BLEU score is a standard metric used to judge similarity between sentences based on the $n$-grams they share.
Levenshtein distance (also known as edit distance) is a standard metric to distance between two strings: the minimum of swap/deletion/insertion operations that are needed to convert one string to another.
To measure similarity between $s_1$ and $s_2$, we use $(\texttt{maxlen}(s_1, s_2) - \texttt{Levenshtein}(s_1, s_2)) / \texttt{maxlen}(s_1, s_2)$/
Using the nltk.translate.bleu\_score module, we define the average of BLEU-1, BLEU-2 and BLEU-3 metrics by passing \texttt{weights=[1/3, 1/3, 1/3]} in the sentence\_bleu function.
For computing Levenshtein Distances, we utilize the python-Levenshtein package's distance function.
As described in the histograms presented in \cref{fig:lev_plots} and \cref{fig:bleu_plots}, we find much higher similarity scores when employing self-prompting. This logically follows the higher alignment values of such runs. More specifically, however, the similarity of the two scores is ultimately more contingent on the verbiage of the output. As indicated in \Cref{fig:methods}, the SP often closely tracks the exact output of the numeric response and simply replaces the numbers with the respective variables/symbolic expressions, and outputs an expression instead of a final number.
While metrically evident in the provided plots, we see that this ``mirroring'' phenomenon occurs frequently with the use of SP, evident through the high density of similarity scores close to 1 in \Cref{fig:lev_plots}.

\subsection{More on Alignment}
\label{sec:more_alignment}

While we find that the use of the alignment prompt is effective in raising both the accuracy and alignment of a symbolic problem, we run a few supplementary experiments to investigate this behavior even further. When giving the model the alignment prompt (see \cref{table:prompt_format}) from the beginning, not simply when the numeric and symbolic outputs do not align, we actually find a decrease in accuracy from the self-prompting + alignment prompt run. CoT accuracy is $62\%$ and vanilla accuracy is $60.9$\%. Similarly, alignment accuracies are $61.5\%$ and $60.4\%$ for CoT and vanilla, respectively.
When evaluating alignment for the base self-prompting run, we find that the model aligns $83.9\%$ when the numeric output is correct, and $29.7\%$ when it is wrong. Such numbers possibly suggest the model's cognizance of whether or not the numeric evaluation was performed correctly; an implicit understanding of mathematical problem solving.

\subsection{Difficulty of Problems}

\begin{figure*}[t!]
    \centering
    \includegraphics[scale=0.4]{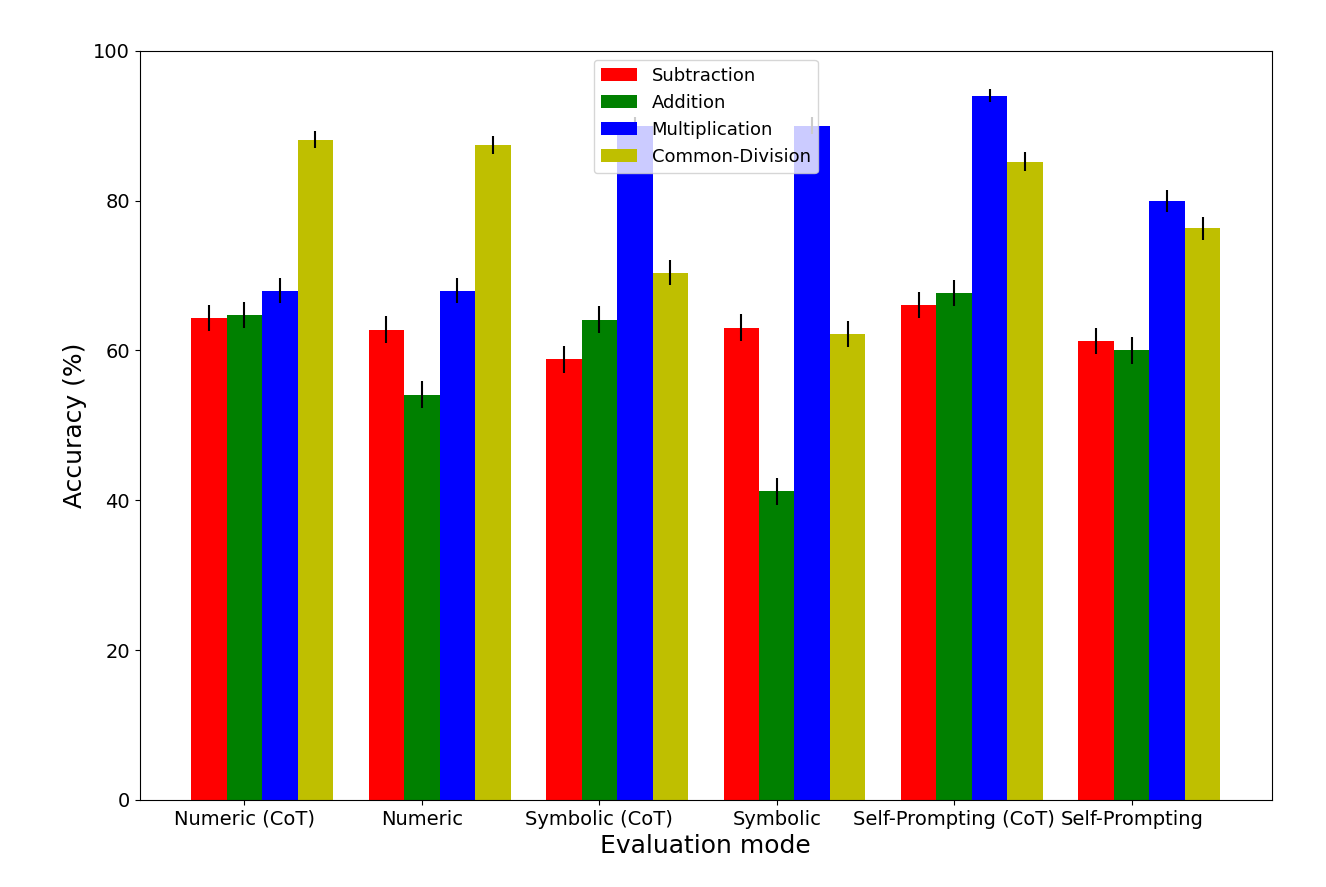}
    \caption{Accuracies of the ``tag'' of the prompt inputted into the model based on the evaluation method of the model. We observe that numeric consistently performs above average at division, symbolic at multiplication, and self-prompting at both. By combining the strengths of both numeric and symbolic evaluation, we see that self-prompting is able to perform as well, if not better than both numeric and symbolic prompting. Furthermore, as with general accuracy CoT also seems to provide boosts to addition accuracies, emphasized especially when comparing symbolic evaluations (Vanilla and CoT).}
    \label{fig:tag_barplot}
\end{figure*}

\begin{table*}[]
\begin{center}
\begin{tabular}{cc|cccc|}
\cline{3-6}
 &  & \multicolumn{4}{c|}{\textbf{Accuracy (\%)}} \\ \cline{2-6} 
\multicolumn{1}{c|}{} & Evaluation & \multicolumn{1}{c|}{Addition} & \multicolumn{1}{c|}{Subtraction} & \multicolumn{1}{c|}{Multiplication} & Division \\ \hline
\multicolumn{1}{|c|}{\multirow{2}{*}{\textbf{Numeric}}} & \textit{CoT} & \multicolumn{1}{c|}{64.7} & \multicolumn{1}{c|}{64.3} & \multicolumn{1}{c|}{68} & 88.1 \\ \cline{2-6} 
\multicolumn{1}{|c|}{} & \textit{Vanilla} & \multicolumn{1}{c|}{54.1} & \multicolumn{1}{c|}{62.8} & \multicolumn{1}{c|}{68} & 87.4 \\ \hline
\multicolumn{1}{|c|}{\multirow{2}{*}{\textbf{\begin{tabular}[c]{@{}c@{}}Symbolic\\ \{w, x, y, z\}\end{tabular}}}} & \textit{CoT} & \multicolumn{1}{c|}{64.1} & \multicolumn{1}{c|}{58.8} & \multicolumn{1}{c|}{90} & 70.4 \\ \cline{2-6} 
\multicolumn{1}{|c|}{} & \textit{Vanill}a & \multicolumn{1}{c|}{41.2} & \multicolumn{1}{c|}{63} & \multicolumn{1}{c|}{90} & 62.2 \\ \hline
\multicolumn{1}{|c|}{\multirow{2}{*}{\textbf{\begin{tabular}[c]{@{}c@{}}Self-prompting\\ \{w, x, y, z\}\end{tabular}}}} & \textit{CoT} & \multicolumn{1}{c|}{67.6} & \multicolumn{1}{c|}{66.1} & \multicolumn{1}{c|}{94} & 85.2 \\ \cline{2-6} 
\multicolumn{1}{|c|}{} & \textit{Vanilla} & \multicolumn{1}{c|}{60} & \multicolumn{1}{c|}{61.3} & \multicolumn{1}{c|}{80} & 73.3 \\ \hline
\end{tabular}
\caption{While the accuracies presented are fairly consistent within each separate evaluation run, we see that there are clear biases in which the model is able to perform certain types of problems better depending on the context of the run. Significantly, it should be noted that the self-prompting is able to employ both the efficiencies of numeric, \textit{and} symbolic runs with the increased alignment.}
\label{table:type_analysis_table}
\end{center}
\end{table*}

We highlight a metric for the difficulty of a problem with respect to the primary operation performed in the answer to the prompt. The SVAMP dataset stores a ``Type'' key that denotes the primary elementary mathematical operation performed to get to the answer (primary in cases where there is more than one operation). We see that when graphing the accuracies of various evaluation methods while isolating the operation of the problem that the numeric and symbolic runs exhibit a somewhat complementary behavior. While numeric does on average better on problems with division, symbolic runs have higher accuracy on multiplication, see \cref{fig:tag_barplot}. \cref{table:type_analysis_table} has breakdowns of the exact accuracies per each tag. 
Interestingly, the self-prompting approach seems to do well on both multiplication and division, and its performance is close to the max of the numeric and symbolic performance for each category, thus hinting to a ``best of both worlds'' phenomenon.

\section{Additional Details}
\label{sec:add_details}

\subsection{Prompt formats}
\label{sec:prompt_format}
In the SVAMP dataset, each problem contains a problem statement and a question. For both raw numeric and symbolic evaluations, we input the problem into the model with the CoT prompt if appropriate. For self-prompting, however, in order to increase alignment between the numeric and symbolic outputs, we add the entire transcript of the numeric evaluation (problem, answer prompting, symbolic problem). A detailed transcript of each of the different prompts and use cases can be found in \cref{table:prompt_format}.
\newcommand{\prompt}[1]{{\color{blue}\texttt{<#1>}}}
\newcommand{\response}[1]{{\color{green!40!black}\texttt{<#1>}}}

\begin{table*}[t!]
    \small
    \begin{center}
\begin{tabular} { | c | p{12cm} | } 
  \hline
  Example
  & 
  \texttt{\small \prompt{Numeric Setup} = "Adam had 5 apples. He ate 2 of them for breakfast." \newline 
  \prompt{Numeric Question} = "How many apples will he have left if he eats 1 more?" \newline 
  \prompt{Symbolic Setup} = "Adam had w apples. He ate x of them for breakfast." \newline 
  \prompt{Symbolic Question} = "How many apples will he have left if he eats y more?"}\\

  \hline

  Prompts &
  \texttt{\small \prompt{CoT Prompt} = "Let's think step by step." \newline 
  \prompt{Numeric Extract Prompt} = "The final answer (only the number) is:" \newline 
  \prompt{Symbolic Extract Prompt} = "The final answer (only the expression in terms \newline {\color{white} xxxxxxxxxxxxxxxxxxxxxxxxxxxx} of given variables) is:" \newline 
  \prompt{Align Prompt} = "Copy the above numeric response word to word but\newline {\color{white} xxxxxxxxxxxxxxxxx} replace numbers with the right symbolic expression."}\\

  \hline

  Numeric &
  \texttt{Q: \prompt{Numeric Setup} \prompt{Numeric Question} \newline 
  A: \prompt{CoT Prompt} 
  \response{Numeric Response} {\color{gray}// language model's verbose response}\newline
  \prompt{Numeric Question} \prompt{Numeric Extract Prompt} \newline
  \response{Numeric Extracted}
  }\\
  
  \hline

  Symbolic &
  \texttt{Q: \prompt{Symbolic Setup} \prompt{Symbolic Question} \newline 
  A: \prompt{CoT Prompt} 
  \response{Symbolic Response} {\color{gray}// language model's verbose response}\newline
  \prompt{Symbolic Question} \prompt{Symbolic Extract Prompt} \newline
  \response{Symbolic Extracted}
  }\\
  
  \hline

  Self-prompt &
  \texttt{Q: \prompt{Numeric Setup} \prompt{Numeric Question} \newline 
  A: \prompt{CoT Prompt} 
  \response{Numeric Response} \newline
  \prompt{Align Prompt} {\color{gray}// [optional] only if alignment fails without it} \newline
  Q: \prompt{Symbolic Setup} \prompt{Symbolic Question} \newline 
  A: \prompt{CoT Prompt} 
  \response{Symbolic Response} \newline
  \prompt{Symbolic Question} \prompt{Symbolic Extract Prompt} \newline
  \response{Symbolic Extracted}
  }\\

  \hline
\end{tabular}
\end{center}
\caption{We present the prompting pipeline for various methods. Prompts in {\color{blue}blue} are the ones we pass to the model, while the text in {\color{green!40!black}green} are the output of the language model. In each of these methods, we include a final filtering step on top of the extracted answers.}
\label{table:prompt_format}
\end{table*}

\subsection{Filtering}
\label{sec:filtering}

Since there is high variability in the LM's outputs, due to the necessity to reason when solving a MWP, we employ several filtering techniques in a \texttt{filter()} function that cleans up the extracted numeric or symbolic output. A few main steps in the filtering pipeline are as follows:
\begin{itemize}
    \item Character replacing
    \begin{itemize}
        \item Dollar signs
        \item Percentages
    \end{itemize}
    \item Cleaning up the output by removing all words besides the expression and/or final number
    \item Addressing cases of outputs such as code or \LaTeX
    \item Isolating the outputted/final expression if the answer is given in terms of an equation (say ``\texttt{z = w + x}'')
\end{itemize}
The detailed (pseudo) code of the function can be found at the end.
\clearpage
\onecolumn
\begin{python}
def filter_symbolic(response):

    response = response.lower()
    response = response.strip('\n')
    print(f"Original Output: {response}")

    # De-latexifying
    response = LatexNodes2Text().latex_to_text(response)
    response = response.replace("$", "")

    # Using * as multiplication operator
    response = response.replace('.', '*')

    # Handling the division symbol
    response = response.replace("%
    response = response.replace('\u00F7', '/')

    # Remove spaces and construct a boolean array denoting whether
    # the character is in the set {'w', 'x', 'y', 'z', '/', '*', '+', '-', '(', ')'}
    math_sym_set = set(['w', 'x', 'y', 'z', '/', '*', '+', '-', '(', ')'] +\
    [str(a) for a in range(10)])

    # Check for "words" that only contain chars from math_sym_set
    response = response.replace("=", " = ")
    words = response.lower().split()
    is_math_sym = np.array([np.all([c in math_sym_set for c in word])*len(word) for word in words])

    # Pick the substring with non-zero entries that has the largest sum,
    # i.e. the largest substring of the original string that is an equation/expression
    idx, len_ = longest_sum(is_math_sym)
    response = ''.join(words[idx:idx+len_])
    print(response)

    # Add multiplication operator * if needed.
    # Logic: If neither of two consecutive characters is an operator
    # then likely a multiplication operator needs to be added between them.
    # Some edges cases like '(p' or 'q)' are handled
    op_set = set(['/', '*', '+', '-'])
    digit_set = set([str(a) for a in range(10)])
    new_response = []
    for i in range(len(response)):
        new_response.append(response[i])
        # Check if '*' needs to be added
        if i < len(response)-1 and response[i] not in op_set and response[i+1] not in op_set:
            # No need to add '*' if the consecutive chars of the type '(p' or 'q)' of '25'
            if (response[i] != '(' and response[i+1] != ')') and (response[i] not in digit_set or response[i+1] not in digit_set):
                new_response.append('*')

    print(f"Final Output: {new_response}")
    return ''.join(new_response)
    return output

def filter_numeric(response):
    output = str(response).replace(",", "")
    output = output.replace("$", "")
    output = output.strip('\n')
    try:
        output = int(re.findall('\d+', output)[0])
    except:
        output = output
    return output
\end{python}
\twocolumn

\end{document}